\documentclass[francais]{gretsi}

\usepackage[utf8]{inputenc}

\usepackage[T1]{fontenc}

\usepackage{amsmath, amssymb, amsfonts}

\usepackage{xcolor}

\usepackage{float}
\usepackage[justification=centering]{caption}
\usepackage{hyperref}
\usepackage{comment} 
\usepackage{multicol}
\usepackage{multirow}
\DeclareMathOperator*{\argmin}{arg\,min}

\let\OLDthebibliography\thebibliography
\renewcommand\thebibliography[1]{
  \OLDthebibliography{#1}
  \setlength{\parskip}{0pt}
  \setlength{\itemsep}{0pt plus 0.3ex}
}

\usepackage[bottom]{footmisc}

\begin{document}


\titre{Détection d'anomalies dans l'espace image ou l'espace latent d'auto-encodeurs par patch pour l'analyse d'images industrielles}



\auteurs{
  \auteur{Nicolas}{Pinon}{nicolas.pinon@creatis.insa-lyon.fr}{1}
  \auteur{Robin}{Trombetta}{robin.trombetta@creatis.insa-lyon.fr}{1}
  \auteur{Carole}{Lartizien}{carole.lartizien@creatis.insa-lyon.fr}{1}
}

\affils{
  \affil{1}{
  Univ Lyon, INSA‐Lyon, Université Claude Bernard Lyon 1, CNRS, Inserm, CREATIS UMR 5220, U1294, F‐69621, LYON
  }
  }


\resume{Nous étudions plusieurs méthodes de détection d'anomalies dans des images couleurs, basées sur des auto-encodeurs par patch.
Nous comparons les performances de trois types de méthodes basées, la première, sur l’erreur entre l'image originale et sa reconstruction, la seconde, sur l’estimation du support de la distribution des images normales dans l'espace latent, et la troisième, sur l’erreur entre l’image originale et une version restaurée de l’image reconstruite.
Ces méthodes sont évaluées sur la base d'images industrielles MVTecAD et comparées à deux méthodes très performantes de l'état de l'art.
}


\abstract{We study several methods for detecting anomalies in color images, constructed on patch-based auto-encoders. 
We compare the performance of three types of methods based, first, on the error between the original image and its reconstruction, second, on the support estimation of the normal image distribution in the latent space, and third, on the error between the original image and a restored version of the reconstructed image.
These methods are evaluated on the industrial image database MVTecAD and compared to two competitive state-of-the-art methods. \\
}

\maketitle


\section{Introduction}

La détection d’anomalie est un paradigme d’apprentissage alternatif aux méthodes supervisées et adapté à des tâches pour lesquelles il est soit trop coûteux d’obtenir des données annotées, soit difficile de caractériser le type d’anomalie rencontré.
Dans ce contexte, on cherche à paramétrer un modèle statistique de représentation de la normalité à partir de données "saines", c'est-à-dire ne contenant pas d'anomalies. En inférence, les échantillons tests s'écartant de la distribution normative sont considérés comme des anomalies (\textit{outlier}). L'architecture classique d'un modèle de détection d’anomalies est constituée de plusieurs modules, un module (facultatif) d’encodage de l’information contenue dans les données d'entrée dans un espace de représentation, suivi d’une modélisation statistique de la distribution normative (modèle génératif) ou du support de cette distribution (modèle discriminatif) dans l’espace de représentation et enfin d’une mesure de distance à la distribution normative qui permet détecter la présence de données \textit{outliers}.
De nombreux développements méthodologiques ont été proposés récemment dans ce domaine \cite{padim, fastflow} avec des applications en vision par ordinateur. Ces développements ont été menés conjointement avec la constitution d'une base de données d'images industrielles, la base MVTecAD \cite{mvtecad}, mise à la disposition de la communauté et permettant une analyse comparative des performances des différents algorithmes de détection d'anomalies.

\noindent Nous proposons d'étudier plusieurs méthodes de détection d'anomalies, basées sur des auto-encodeurs par patch (ou imagette), pour des tâches de détection d'anomalies dans des images couleurs.
L'intérêt des approches par patch réside dans le travail à petite échelle qui permet 1) de mettre l'accent sur de petites structures, 2) d'augmenter la taille des bases d'entrainement (grand nombre de patchs disponibles) ou encore 3) d'entrainer des modèles plus légers 
au prix cependant d'une perte de contexte global. 

\noindent Nous comparons les performances de trois types de méthodes. Le premier type de méthodes consiste à mesurer l'erreur entre l'image originale donnée à l'entrée d'un auto-encodeur et sa reconstruction à la sortie. L'hypothèse sous-jacente est qu'un auto-encodeur, entrainé sur des images 'normales' (i.e. sans anomalie), reconstruira moins bien une image présentant un défaut à l'inférence. Une deuxième famille de méthodes essaie d'estimer directement le support de la distribution des images normales dans l'espace latent continu et compressé de l'auto-encodeur. Enfin, une troisième famille de méthodes s'appuie sur la discrétisation de cet espace latent pour apprendre un modèle auto-régressif permettant de "restaurer" les images déviant de la normalité, c'est-à-dire les rapprocher d'images normatives avant de les reconstruire.
L'originalité de nos contributions est la suivante : jusqu'ici, à notre connaissance, tous les modèles évalués sur MVTec AD reposant sur la modélisation de l'espace latent d'images nécessitent un réseau extracteur de caractéristiques visuelles pré-entrainé sur une très large base de données telle que ImageNet (ce qui peut s'avérer long, couteux, et peu transférable à d'autres applications). D'autre part, nous nous intéressons spécifiquement à des méthodes par patch et considérons des espaces latents continu ou discrétisé. Enfin, nous comparons ces modèles à deux méthodes très performantes de l'état de l'art, entrainées dans les mêmes conditions que nos auto-encodeurs par patch, les méthodes PaDiM \cite{padim} et Fastflow \cite{fastflow}, sur deux types d'objets texturés de la base MVTecAD : des photos de planches de bois (\textit{wood}) et de tissus d'ameublement unis (\textit{carpet}).

\vspace{-11px}
\section{Méthodes}

\begin{figure*}[t]
  \centering
  \includegraphics[scale=0.40]{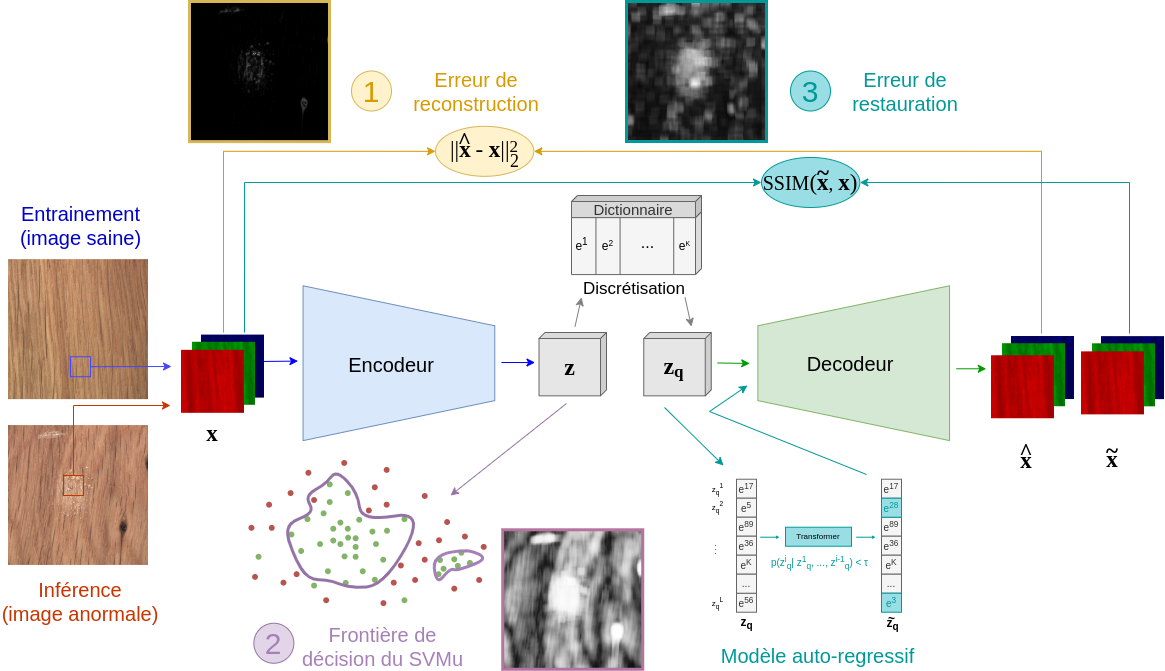}
  \caption{\small Schéma synthétique des méthodes présentées. L'auto-encodeur est entrainé avec des patchs d'images normales. A l'inférence, on présente des patchs extraits d'images anormales. Les méthodes continues ne tiennent pas compte du processus de discrétisation.}
    \label{fig:method}
\end{figure*}

\vspace{-5px}
\subsection{Apprentissage de représentation}



\noindent \textbf{Apprentissage de représentation continue} \\
\noindent Un auto-encodeur est d'abord entraîné à reconstruire des patchs (i.e. imagettes) d'images "normales" avec une fonction de coût définie comme :
\vspace{-5px}
$$L_{\scriptscriptstyle AE}(\mathbf{x}) =  ||\mathbf{x} -\mathbf{\hat{x}}||_2^2 \qquad \text{avec} \quad \mathbf{\hat{x}} = D(z) = D(E(x))$$
\vspace{-18px}
\par\noindent\ignorespaces $\mathbf{x}$ un patch d'image RVB et $\mathbf{\hat{x}}$ sa reconstruction après décodage $D$ de sa représentation latente $\mathbf{z}$, obtenue par encodage $E$.

\noindent L'utilisation de patchs permet d'augmenter artificiellement la taille de la base de données, et donc d'introduire un effet de régularisation, qui peut limiter le sur-apprentissage. Cet apprentissage est dit "continu" car il n'y a aucune restriction sur les valeurs que peut prendre $\mathbf{z}$.


\noindent \textbf{Apprentissage de représentation discrétisée} \\
\noindent Une alternative à l'apprentissage de représentation continue consiste à forcer les représentations latentes à prendre des valeurs discrètes \cite{vqvae}. Ces valeurs sont prises dans un dictionnaire appris au cours de l'entrainement $\mathcal{E}_K = \{e^1, ..., e^K\}$ de taille $K$, chaque coordonnée $z^j$ de la représentation latente du patch $\mathbf{z}$ est remplacée par le vecteur du dictionnaire $e^k$ le plus proche ($k = \argmin_l{||z^j - e^l||_2^2}$), donnant une représentation discretisée $\mathbf{z_q}$. La fonction de coût de ce modèle s'exprime comme :
\vspace{-5px}
$$L_{\scriptscriptstyle AE \, discret.}(\mathbf{x}) =  ||\mathbf{x} - \mathbf{\hat{x}}||_2^2 + ||sg[\mathbf{z}] - \mathbf{z_q}||_2^2 + \beta||\mathbf{z} - sg[\mathbf{z_q}]||_2^2$$

\vspace{-3px}
\par\noindent\ignorespaces Avec $\mathbf{z_q} = (z_q^1, ..., z_q^L)$,  $\mathbf{z} = E(\mathbf{x}) = (z^1, ..., z^L)$ et $\mathbf{\hat{x}} = D(\mathbf{z_q})$, $L$ la dimension spatiale de l'espace latent et $sg[\cdot]$ désigne l'opérateur d'arrêt de gradient.
\noindent L'intérêt d'un espace latent discretisé peut être d'utiliser des modèles auto-regressifs pour la prédiction de certaines coordonées de $\mathbf{z_q}$ à partir d'autres, ou dans le but de structurer différemment l'espace latent.
\vspace{-11px}
\subsection{Détection d'anomalies}

\noindent \textbf{Erreur de reconstruction} \\
\noindent La méthode standard pour la détection d'anomalie consiste à estimer, à l'inférence, l'erreur $||\mathbf{x} -\mathbf{\hat{x}}||_2^2$ utilisée aussi pendant l'entrainement. 
Cette méthode est applicable dans le cas de l'apprentissage de représentation continue ou discrète. Dans ce travail, seule l'erreur du pixel central est utilisée pour obtenir les cartes d'anomalies.

\noindent \textbf{SVM uniclasse} \\
\noindent Une méthode alternative à l'erreur de reconstruction,  proposée dans \cite{midl},
consiste à estimer le support de la distribution de probabilité des patchs normaux dans l'espace latent.
Pour cela, on extrait les représentations latentes $\mathbf{z}_i$ des patchs d'entrainement $\mathbf{x}_i$, et on construit, à l'aide d'un séparateur à vaste marge (SVM) uniclasse, 
une fonction $f$, positive sur le domaine des $\mathbf{z}_i$ et négative à l'extérieur.
Cette fonction est construite par projection des $\mathbf{z}_i$ dans un espace de plus haute dimension à l'aide d'une transformation $\mathbf{\phi}(.)$ où ils sont séparés de l'origine par un hyperplan. 
Un score d'anomalie correspondant à la distance à l'hyperplan peut ensuite être attribué au pixel central de chaque patch.

\noindent \textbf{Erreur de restauration}\\
\noindent Une méthode hybride, consiste à modéliser la distribution dans l'espace latent, dans le but de corriger (re-échantilloner) les anomalies, puis utiliser le décodeur pour repasser dans l'espace image et comparer l'image corrigée (aussi appelée restaurée) à l'image originale. Ceci peut permetre de limiter les faux positifs si le patch restauré est proche du patch initial dans l'espace image, ce qui ne serait pas possible en évaluant directement la probabilité dans l'espace latent.
L'auto-encodeur discretisé peut être associé à un modèle auto-régressif 
pour modéliser la distribution\footnote{En réalité ce sont les probabilités des indices $k$ des $e^k \in \mathcal{E}$ qui sont appris et non pas directement les probabilités des $e^k$. Aussi, les $e^k$ et $z^j$ ne sont pas scalaires mais de dimension $d$.} $p(z^j_q | z^1_q,...,z^{j-1}_q)$ des coordonées du vecteur $\mathbf{z_q}$ qui prennent valeur dans $\mathcal{E}$. Durant la phase d'inférence, si la probabilité d'un élément de la séquence est inférieure à un seuil $\tau$ préalablement fixé, alors cet élément est ré-échantillonné. Ce vecteur restauré $\mathbf{\Tilde{z}}_{\mathbf{q}}$ est décodé pour obtenir un patch restauré $\mathbf{\Tilde{x}} = D(\mathbf{\Tilde{z}}_{\mathbf{q}})$. L'erreur de restauration est ensuite calculée comme $SSIM(\mathbf{x}, \mathbf{\Tilde{x}})$ avec $SSIM$ la \textit{Structural SIMilarity}. 

\vspace{-11px}
\section{Expériences}
\vspace{-5px}
\subsection{Base de données d'images industrielles}

La base de donnée utilisée est la base MVTecAD \cite{mvtecad}. Cette base d'images industrielles contient 5354 images, divisées en 15 types : 10 images d'objets et 5 de textures. On se focalisera dans ce travail sur deux sous-catégories d'images de texture, \textit{wood} et \textit{carpet}.
La catégorie \textit{wood} contient des photos de planches de bois, réparties en 247 images d'entrainement (normales), 19 images de test normales et 60 images de tests contenant 5 types possibles de défauts. 
La catégorie \textit{carpet} contient des photos de tissus d'ameublement unis (i.e. sans motifs), réparties en 280 images d'entrainement, 28 images de test normales et 89 images de test défectueuses, avec 5 types de défauts. 
Les images de la catégorie \textit{wood} et \textit{carpet} sont de résolution 1024 par 1024 pixels.
\vspace{-11px}
\subsection{Hyperparamètres des méthodes}
 \begin{figure}[b]
   \centering
   \includegraphics[scale=0.1]{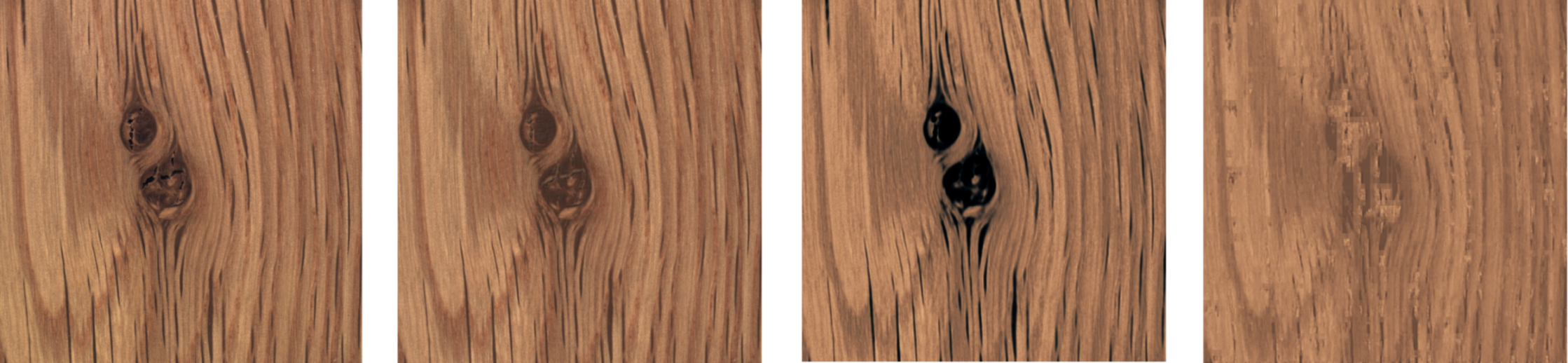}
   \caption{\small Image originale (gauche), reconstruction par AE (centre gauche), reconstruction par AE discretisé (centre droit) et restauration par AE discretisé (droite)}
     \label{fig:comp}
 \end{figure}

L'encodeur est composé de 4 blocs de convolution avec des noyaux de taille $(5, 5)$, $(3, 3)$, $(3, 3)$ et $(3, 3)$, \textit{stride} de respectivement $(1, 1)$, $(1, 1)$, $(3, 3)$ et $(1, 1)$ (donnant une dimension spatiale d'espace latent $L=$ 17$\times$17), nombre de filtres respectivement de $3$, $4$, $12$ et $16$ (donnant une dimension de canaux d'espace latent $d=$ 16), pas de \textit{padding} et une activation GeLu. Chaque bloc est suivi d'un bloc de \textit{batch normalization}.
Le décodeur est construit comme le symétrique de l'encodeur. La taille du dictionnaire $K$ utilisé pour la discrétisation est fixée à 1024.
L'auto-encodeur est entraîné en extrayant 1000 patchs (concaténation RVB) par image de taille 63$\times$63$\times$3, c'est-à-dire 247 000 patchs pour \textit{wood} et 280 000 pour \textit{carpet}.
Nous avons utilisé l'optimiseur Adam sur 10 époques, avec une taille de \textit{batch} de 100. 
Pour le SVM uniclasse, nous avons fixé $\nu = 0.03$ et utilisé un noyau gaussien dont l'hyperparamètre $\frac{1}{\gamma}$ a été fixé comme le produit de la variance et de la dimension des $\mathbf{z}_i$.
L'architecture du Transformer contient 8 blocs de décodeur à 8 têtes avec une dimension de représentation de 256 et une taille de 512 pour les couches denses intermédiaires. Le seuil de ré-échantillonnage $\tau$ est fixé à 5\%. La taille de fenêtre pour le calcul de la \textit{SSIM} est de 63 par 63.
\vspace{-11px}
\subsection{Comparaison à l'état de l'art}

Nous comparons nos modèles avec deux méthodes de l'état de l'art parmi les plus performantes pour la détection d'anomalie sur MVTecAD : PaDiM \cite{padim} et FastFlow \cite{fastflow}. Ces deux modèles reposent sur l'extraction de caractéristiques des images par un réseau de convolutions (ici un ResNet50) et la modélisation de leur distribution par des mixtures de gaussiennes multivariées pour le premier, et un modèle de \textit{Normalizing Flow} pour le second. L'implémentation des modèles est faite grâce à la bibliothèque Anomalib \cite{anomalib}. 
\vspace{-11px}
\subsection{Métriques}

\begin{figure*}[h!]
  \centering
  \includegraphics[scale=0.22]{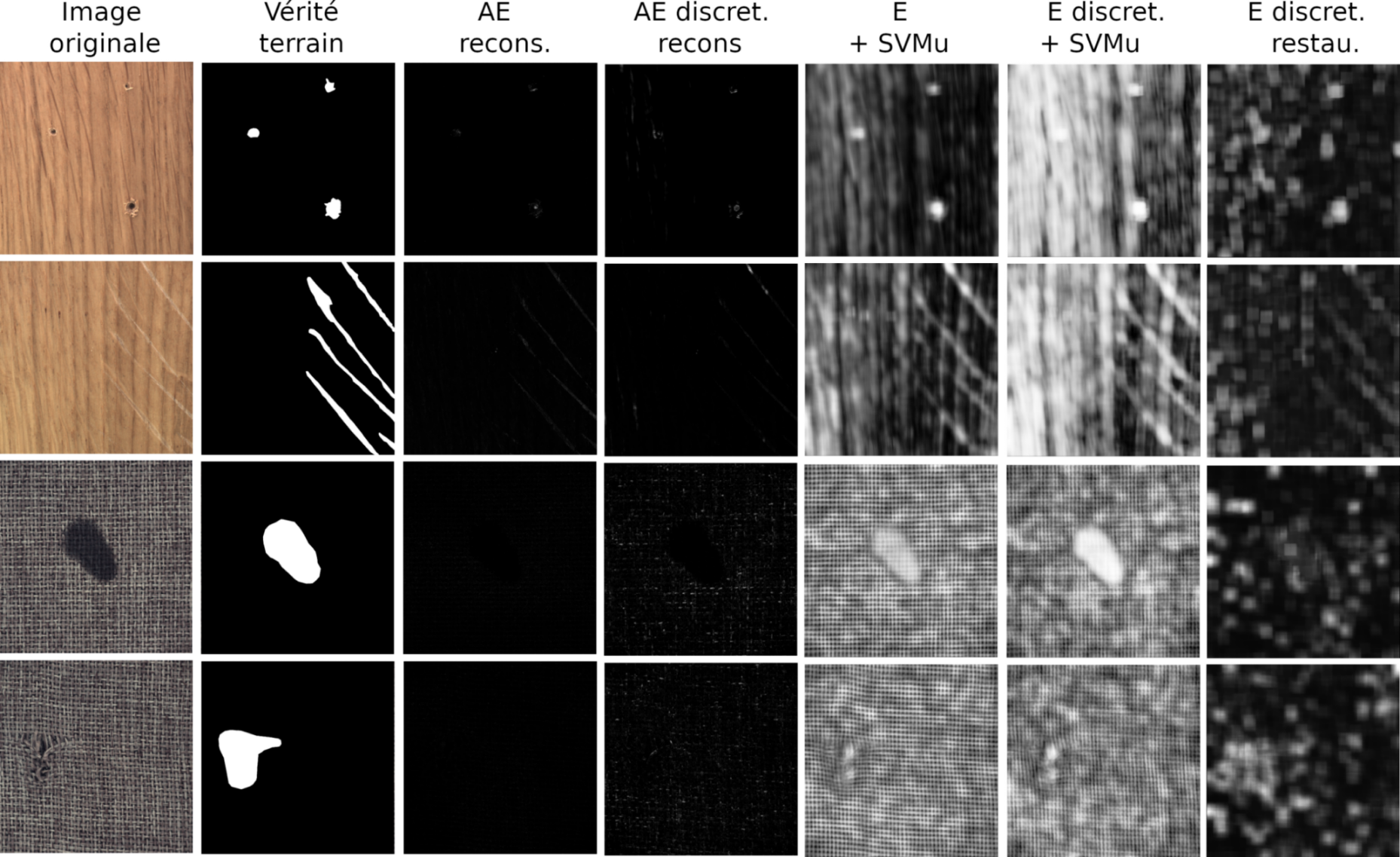}
  \caption{\small Comparaison des méthodes présentées sur 4 examples (2 \textit{wood} et 2 \textit{carpet})}
    \label{fig:results}
\end{figure*}

Les différentes méthodes sont évaluées à l'échelle du pixel grâce à des métriques classiques : l'aire sous la courbe ROC (AUROC), ainsi que l'aire sous la courbe précision-rappel (AUPRC), moins sensible au déséquilibre de classe et qui n'utilise pas le taux de faux positifs. À l'instar de \cite{mvtecad}, nous étudions également l'aire sous la courbe de \textit{Per Region Overlap} ou chevauchement par région (AUPRO), qui mesure la moyenne des sensibilités calculées par anomalie (i. e. zone défectueuse de l'image), ce qui permet de ne pas favoriser la segmentation des grandes anomalies par rapport aux plus petites, un biais très présent avec la mesure d'AUROC.
Nous nous intéressons aussi aux aires sous les courbes ROC et PRO correspondant à un taux de faux positifs inférieur à 30\% \footnote{Il est important de noter que le score d'un classifieur aléatoire pour AUPRO 30 et AUROC 30 est de 0.15. Pour l'AUPRC, ce score serait de 0.02 pour \textit{carpet} et 0.04 pour \textit{wood}.}. En effet, les cartes d'anomalies comportant plus de 30\% de faux positifs peuvent être considérées comme dégénérées et sans intérêt dans le cadre d'anomalies ne couvrant qu'une zone très faible des images (3.9\% pour le jeu de test de \textit{wood} et 1.6\% pour \textit{carpet}).

\begin{table}[]
\centering
\footnotesize
\caption{\small Performances des modèles étudiés sur \textit{wood} et \textit{carpet}.\label{table:results}}
\begin{tabular}{|c|c|c|c|c|c|c|}
\hline
\multirow{ 3}{*}{MVTecAD} & \multicolumn{ 2}{c|}{AUROC} & \multicolumn{2}{c|}{\multirow{2}{*}{AUPRC}} & \multicolumn{ 2}{c|}{AUPRO} \\ 
 & \multicolumn{ 2}{c|}{AUROC 30} & \multicolumn{2}{c|}{} & \multicolumn{ 2}{c|}{AUPRO 30} \\ \cline{ 2- 7}
 & \textit{carpet} & \textit{wood} & \textit{carpet} & \textit{wood} & \textit{carpet} & \textit{wood} \\ \hline
\multicolumn{ 1}{|c|}{AE} & 0.51 & 0.70 & \multirow{2}{*}{0.02} & \multirow{2}{*}{0.29} & 0.57 & 0.75 \\
\multicolumn{ 1}{|c|}{recons.} & 0.17 & 0.45 &  &  & 0.25 & 0.54 \\ \hline
\multicolumn{ 1}{|c|}{AE discret.} & 0.51 & 0.71 & \multirow{2}{*}{0.02} & \multirow{2}{*}{0.26} & 0.51 & 0.55 \\ 
\multicolumn{ 1}{|c|}{recons.} & 0.16 & 0.47 &  &  & 0.16 & 0.26 \\ \hline
\multicolumn{ 1}{|c|}{E} & 0.47 & 0.80 & \multirow{2}{*}{0.02} & \multirow{2}{*}{0.35} & 0.47 & 0.82 \\
\multicolumn{ 1}{|c|}{+ SVMu} & 0.12 & 0.57 &  &  & 0.14 & 0.62 \\ \hline
\multicolumn{ 1}{|c|}{E discret.} & 0.54 & 0.75 & \multirow{2}{*}{0.02} & \multirow{2}{*}{0.29} & 0.57 & 0.78 \\
\multicolumn{ 1}{|c|}{+ SVMu} & 0.15 & 0.50 &  &  & 0.21 & 0.54 \\ \hline
\multicolumn{ 1}{|c|}{AE discret.} & 0.79 & 0.76 & \multirow{2}{*}{0.05} & \multirow{2}{*}{0.27} & 0.76 & 0.83 \\
\multicolumn{ 1}{|c|}{restau.} & 0.45 & 0.56 &  &  & 0.42 & 0.63 \\ \hline
\multirow{2}{*}{PaDiM} & 0.75 & 0.85 & \multirow{2}{*}{0.14} & \multirow{2}{*}{0.32} & 0.69 & 0.83 \\
 & 0.41 & 0.61 & \multicolumn{ 1}{c|}{} &  & 0.34 & 0.59 \\ \hline
\multirow{2}{*}{FastFlow} & 0.83 & 0.85 & \multirow{2}{*}{0.24} & \multirow{2}{*}{0.41} & 0.85 & 0.90 \\
 & 0.59 & 0.65 & \multicolumn{ 1}{c|}{} &  & 0.60 & 0.72 \\ \hline
\end{tabular}

\end{table}
\vspace{-11px}
\section{Résultats et discussion}

Le tableau \ref{table:results} évalue les métriques présentées pour l'erreur de reconstruction de l'auto-encodeur et de l'auto-encodeur discrétisé (AE recons. et AE discret. recons.), le SVM uniclasse après projection de l'encodeur ou encodeur discrétisé (E + SVMu ou E discret. + SVMu), la $SSIM$ entre la restauration et l'image originale (AE discret. restau.), les deux méthodes concurrentes PaDiM et FastFlow, le tout sur les jeux de test de \textit{carpet} et \textit{wood}. La figure \ref{fig:comp} présente les différences entre la restauration et les reconstructions continue et discrète. La figure \ref{fig:results} présente des exemples de cartes de scores d'anomalies obtenues par nos différentes méthodes.


\noindent 
Les performances de FastFlow sont supérieures quelle que soit la métrique ou le type de texture évalué. En fonction de la métrique évaluée, PaDiM ou AE discret. restau. sont fréquemment 2ème ou 3ème. Les méthodes E + SVMu et AE discret. restau. obtiennent des performances en AUPRO très proches de FastFlow sur \textit{wood}. Sur \textit{carpet}, seul AE discret. restau. conserve de bonnes performances parmi nos modèles.


\noindent
Une étude plus approfondie montre une grande disparité de performances en fonction du type d'anomalie à détecter sur les images. Par exemple, E + SVMu obtient une AUPRO de 0.49 seulement sur les images de \textit{wood} contenant une tache de liquide, alors qu'il détecte les trous avec une AUPRO de 0.94. Les modèles semblent meilleurs lorsque l'anomalie à segmenter a un contraste différent des motifs normaux. Ce phénomène pourrait expliquer la chute de performance sur \textit{carpet}, en particulier sur les anomalies de forme comme les incisions ou celles qui ne modifient pas beaucoup la couleur comme avec la contamination par éclat métallique.

\noindent
Un avantage important de l'approche par patch est la légèreté des modèles. La taille de l'auto-encodeur est de seulement quelques milliers de paramètres là où des modèles reposant sur de grands extracteurs de caractéristiques contiennent plusieurs dizaines voire centaines de millions de paramètres. Ils demandent aussi généralement davantage de temps et de ressources pour être entraînés. Un désavantage de l'approche par patch se trouve dans le temps d'inférence, généralement plus long car l'obtention d'une carte de score complète nécessite de nombreux passages dans le modèle.

\noindent 
Dans \cite{mvtecad}, un auto-encodeur entrainé sur 1/4 de l'image et utilisant l'erreur de reconstruction à l'inférence obtient une AUROC 30 de 0.29 sur \textit{carpet} et 0.42 sur \textit{wood}, une AUPRO 30 de 0.31 sur \textit{carpet} et 0.52 sur \textit{wood} et enfin une AUPRC de 0.04 sur \textit{carpet} et 0.20 sur \textit{wood}. Ces performances, à comparer avec AE recons. (première ligne de la table \ref{table:results}) qui procède de manière très similaire mais sur des imagettes, n'indique pas clairement une supériorité de l'une ou l'autre échelle de travail lorsqu'on s'intéresse seulement à l'erreur de reconstruction.

\noindent Dans leurs articles originaux respectifs, PaDiM et Fastflow rapportent des AUROC de 0.94 et 0.97 sur \textit{wood} et 0.99 et 0.99 sur \textit{carpet}. Ces valeurs, bien supérieures à celles présentées dans la table \ref{table:results}, ne sont pas comparables car elles sont obtenues à l'aide d'un pré-entrainement du réseau d'apprentissage de représentation sur ImageNet (1.5 millions d'images). 

\noindent Ce travail a permis de positionner des méthodes originales de détection d’anomalies basées sur des auto-encodeurs par patch par rapport à l’état de l’art, sur une base de données de référence, la base MVTecAD. Des perspectives à ce travail visent à mieux analyser les limites de nos modèles (en particulier en fonction du type d’anomalies), à l’évaluer sur d’autres classes d’objet de MVTecAD et à explorer leur performance pour des tâches de détection d’anomalies en imagerie médicale.

\vspace{-11px}
\section{Remerciements}

Ces travaux ont bénéficié d’un accès aux moyens de calcul de l’IDRIS, attribué par GENCI (dossier 2022-AD011012813R1). Ils ont été partiellement financés par le projet
ANR-11-INBS-0006 (FLI) et le projet DAIAA financé par la Fédération d'Informatique Lyonnaise (FIL).
\vspace{-12px}
\bibliography{biblio_et_al}

\begin{thebibliography}{1}
\expandafter\ifx\csname fonteauteurs\endcsname\relax
\def\fonteauteurs{\scshape}\fi

\bibitem{mvtecad}
Paul \bgroup\fonteauteurs\bgroup Bergmann\egroup\egroup{} \emph{et~al.} :
\newblock {The MVTec Anomaly Detection Dataset: A Comprehensive Real-World
  Dataset for Unsupervised Anomaly Detection}.
\newblock {\em International Journal of Computer Vision},
  129(4)\string:\penalty500\relax 1038--1059, Apr 2021.

\bibitem{padim}
Thomas \bgroup\fonteauteurs\bgroup Defard\egroup\egroup{} \emph{et~al.} :
\newblock {PaDiM: A Patch Distribution Modeling Framework for Anomaly Detection
  and Localization}.
\newblock \emph{In} {\em Pattern Recognition. ICPR}, 2021.

\bibitem{midl}
Nicolas \bgroup\fonteauteurs\bgroup Pinon\egroup\egroup{} \emph{et~al.} :
\newblock {One-Class SVM on siamese neural network latent space for
  Unsupervised Anomaly Detection on brain MRI White Matter Hyperintensities}.
\newblock \emph{In} {\em MIDL}, volume in press, 2023.

\bibitem{anomalib}
Akcay \bgroup\fonteauteurs\bgroup Samet\egroup\egroup{} \emph{et~al.} :
\newblock Anomalib: A deep learning library for anomaly detection, 2022.

\bibitem{vqvae}
Aaron van~den \bgroup\fonteauteurs\bgroup Oord\egroup\egroup{} \emph{et~al.} :
\newblock {Neural Discrete Representation Learning}.
\newblock \emph{In} {\em Advances in Neural Information Processing Systems},
  volume~30, 2017.

\bibitem{fastflow}
Jiawei \bgroup\fonteauteurs\bgroup Yu\egroup\egroup{} \emph{et~al.} :
\newblock {FastFlow: Unsupervised Anomaly Detection and Localization via 2D
  Normalizing Flows}.
\newblock {\em CoRR}, 2021.

\end{thebibliography}


\end{document}